\documentclass[]{article}
\usepackage{import}
\import{./}{preamble.sty}

\title{Full Title: Token-UNet: A New Case for Transformers Integration in Efficient and Interpretable 3D UNets for Brain Imaging Segmentation\\ Short Title: Token-UNet: revisiting UNets and Transformers}
\author{Louis Fabrice Tshimanga, Andrea Zanola, Federico Del Pup, Manfredo Atzori}

\begin{document}

\maketitle
\doublespacing

\begin{abstract}
    We present Token-UNet, adopting the TokenLearner and TokenFuser modules to encase Transformers into UNets.
While Transformers have enabled global interactions among input elements in medical imaging, current computational challenges hinder their deployment on common hardware. 
Models like (Swin)UNETR adapt the UNet architecture by incorporating (Swin)Transformer encoders, which process tokens that each represent small subvolumes ($8^3$ voxels) of the input.

The Transformer attention mechanism scales quadratically with the number of tokens, which is tied to the cubic scaling of 3D input resolution.

This work reconsiders the role of convolution and attention, introducing Token-UNets, a family of 3D segmentation models that can operate in constrained computational environments and time frames.

To mitigate computational demands, our approach maintains the convolutional encoder of UNet-like models, and applies TokenLearner to 3D feature maps. 
This module pools a preset number of tokens from local and global structures.

Our results show this tokenization effectively encodes task-relevant information, yielding naturally interpretable attention maps. 
The memory footprint, computation times at inference, and parameter counts of our heaviest model are reduced to 33\%, 10\%, and 35\% of the SwinUNETR values, with better average performance (86.75\% $\pm 0.19\%$ Dice score for SwinUNETR vs our 87.21\% $\pm 0.35\%$).

This work opens the way to more efficient trainings in contexts with limited computational resources, such as 3D medical imaging.
Easing model optimization, fine-tuning, and transfer-learning in limited hardware settings can accelerate and diversify the development of approaches, for the benefit of the research community.
\end{abstract}

\newpage

\section{Author Summary}
Artificial Intelligence (AI) can assist doctors in identifying brain tumors from MRI scans, but the most advanced AI models ("Transformers") are computationally expensive. They often require resources unavailable to most hospitals and medical research labs. This hardware barrier prevents many institutions from developing or using the best available tools for patient care.
In this study, we developed "Token-UNet," a new image segmentation model designed to run efficiently on standard researchers' hardware. By repurposing a technique that compresses 3D images into a small set of informative "tokens" before processing them, we reduced the memory and speed requirements by nearly 90\% compared to current leading models. Despite being much smaller and faster, our model identified tumor regions with the same accuracy as heavier alternatives. Additionally, our method produces visual maps that show which parts of the brain the model is focusing on, helping doctors trust and understand the predictions. This work demonstrates that high-performance medical AI does not require elite hardware, making these tools more accessible to the global medical community.

\section{Introduction}
Discerning the types of healthy and pathological tissues in a possibly affected organ is a complex task.
It requires knowledge of typical patterns, location specifics, variability across large control populations, as well as common or unique forms that signal health problems.
We introduce Token-UNet and apply it to the problem of brain tumor segmentation.
The data in this task are multimodal, morphologically complex, large in memory space, scarce in labels, thus presenting many of the hurdles in bioimaging.
Token-UNet is a convolutional and attentional neural network.
It respectively addresses local and global pattern recognition, and reaches top-level accuracy, high flexibility, and low time and compute costs compared to state-of-the-art (SOTA) architectures for 3D bioimage segmentation.
Deep neural networks have long been dominating the tasks of classification and segmentation in medical imaging~\cite{hung2023deep}, as well as the prediction of diagnosis or hospital admission based on free text reports~\cite{locke2021natural}, to name a few~\cite{acosta2022multimodal}~\cite{duan2024deep}.
In this context, the Brain Tumor Segmentation (BraTS) Challenge~\cite{menze2014multimodal} has been pivotal for measuring the impact and transferability of deep learning techniques in the medical field.
The challenge consists in correctly classifying tumor and lesion tissues in 3D multimodal MRI scans of subjects affected by glioma.
Winning algorithms of the BraTS challenge are often successful in other tissue and organ segmentation challenges~\cite{isensee2021nnu}, with datasets coming from different systems of the human body.
For this reason, the BraTS leaderboard is informative of medical deep learning trends.
One such trend is the dominance of convolutional models derived from UNet~\cite{ronneberger2015u} across other medical imaging applications~\cite{siddique2021u}.
A second major trend is the progressive introduction of Transformer \cite{vaswani2017attention} architectures, with their computation-heavy, expressive attention layers.
Both convolutions and the attention mechanism have specific feature detection properties of relevance to the BraTS challenge.
Tissues tied to tumor growth have geometrical, physical and physiological properties that set them apart from the surrounding healthy tissues.
Structures appear with different contrasts across scan modalities, in varying shapes and sizes depending on subject and tumor growth stage.
Recognizing such features requires a notion of the expected variability in healthy tissues with regards to the many anatomical structures, and a characterization of how the unhealthy tissues themselves can appear.
Convolutional models excel in detecting learned local patterns, regardless of their position in the input image.
However, features are aggregated only locally, thus long-range correlations can be missed.
Attention models like Transformer, instead, evaluate all pairwise interactions between input tokens regardless of distance, which is computationally burdensome.
Moreover, they need to learn spatial biases from scratch.
The complementarity between convolution and attention has been presented as motivation for hybrid models such as SwinUNETR~\cite{hatamizadeh2021swin}, where attention encoders and convolutional decoders are concatenated in a UNet fashion.
Despite the effectiveness of SwinUNETR, several research questions remain unanswered.
It is unclear if the performance gains over fully-convolutional models are consistent across the board of 3D bioimages~\cite{isensee2024nnu}.
If so, it is yet undetermined how much these improvements are directly dependent on the attentional nature of Transformer encoders, rather than parameter counts, number of operations, and other differences.
Drivers of performance and trade-offs are relatively unexplored, in favor of global metric optimization regardless of costs.
Meanwhile, the introduction of Transformers as encoders increases the computational requirements of both training and inference.
Consequently, both testing theories on legacy models and developing new algorithms from acquired knowledge become more expensive in terms of time, energy, economic resources.
As a step in new directions, in this paper we propose a novel integration of convolutions and attention, encasing a small Transformer between the UNet encoder and decoder, similarly to TransUNet \cite{chen2024transunet}.
However, we depart from the straightforward tokenization of inputs or feature maps, and resort to TokenLearner and TokenFuser \cite{ryoo2021tokenlearner}.
By means of these modules, Token-UNet cuts the time and memory requirements for a Transformer to process a 3D image.
TokenLearner and TokenFuser fix the number of tokens processed by the Transformer, decoupling it from the large input size typical of the domain.
Moreover, they add easily inspectable attention maps that open a window of interpretability onto the model decision process.
We statistically estimate performance gains of each modification of a template UNet architecture, until completing a Token-UNet model.
As a result, we start challenging the idea that Transformer encoders or large parameter counts are the most effective and efficient means to achieve high segmentation accuracy.
Evaluating the effect of attentional encoders and shaping new blends of convolutional and attentional layers could improve and democratize both pretraining and fine-tuning of models, allowing better suited applications in biomedical imaging domains, such as image classification and segmentation for diagnostic purposes.

\subsection{Related Works}
UNet is a Convolutional Neural Network (CNN) autoencoder, named after its "U"-shaped structure.
The descending curve of the "U" shape refers to the encoder part of the network, that processes progressively more, lower-resolution, and longer-range features.
Convolutional layers compare neighborhoods of voxels to specific intensity patterns (kernels), thus encoding how small structures are distributed in space.
By reducing the resolution and size of data along the encoder (downsampling), kernels of a fixed size can uncover patterns over larger neighborhoods of the input.
The ascending curve of the "U" shape refers to the decoder.
Each layers of the decoder combines the output from the previous decoder layer with the output from the corresponding encoder layer, which has the same resolution and size, and then upsamples the result.
Each decoder layer thus combines semantically enriched data from the previous decoder layer, with geometrically correct data from the corresponding encoder layer, more similar to the original input.
Many architectures build on these effective principles from UNet, focusing on several modifications.
In contrast, the nnU-Net ("No new net") framework~\cite{isensee2021nnu}~\cite{isensee2024nnu} focuses on inferring proper training hyperparameters from a dataset.
Once the dataset footprint is determined, a tailored sized UNet is trained, without over-engineering new architectures with the risk of overfitting a dataset.
The framework has been applied successfully to the BraTS challenge, among an array comprising 23 datasets and 53 segmentation tasks of varying shapes (2D and 3D), object scales (cells to organs) and acquisition modalities.
Nonetheless, the widespread success of Transformer models outside language-based domains (e.g. Vision Transformers (ViTs)~\cite{dosovitskiy2020image}) has prompted experiments with new architectures in biomedicine~\cite{khan2022transformers}~\cite{parvaiz2023vision}, including the BraTS challenge.
One appealing feature of Transformers is the all-to-all information exchange between input elements, whereas CNN kernels are locally constrained and may reach a global range only when stacked.
The classification of a voxel of brain tissue may depend on its belonging to complex structures that interact at multiple scales, and the Transformer is assumed to express and encode such relationships better than convolutional models.
In order to employ Transformer encoders in 3D vision, the input is divided into fixed size, non-overlapping cubic patches of voxels.
Each patch is flattened into an array of voxels and linearly projected into a vector of dimension $d$, called token embedding.
This tokenization process can be applied either to intermediate convolutional feature maps \cite{chen2024transunet}, or to the original input scan \cite{hatamizadeh2022unetr}, bypassing the convolutional encoder typical of UNet.
Token embeddings pass through Transformer blocks, where they are further projected and interpolated with one another, transcending the distance of the voxel patches they represent.
The main strength of the Transformer block is also a potential hindrance.
The Self-Attention operator computes all pairwise comparisons between token embeddings.
In particular, a Self-Attention head computation on a set $X$ of $N$ tokens of dimension $d$ is defined as:
\begin{equation}
    Z=\text{SA}(X)=\text{softmax}\left( \frac{XW_QW_K^TX^T}{\sqrt{d}} \right)XW_V, \quad \text{with projection matrices }W_Q, W_K, W_V \in \mathbb{R}^{d\times d},
\end{equation}
and it has $O(N^2)$ computational complexity~\cite{keles2023computational}.
For 3D images, the number of tokens grows with the cube of patch resolution: doubling resolution or side incurs in an 8-fold ($2^3$) increase in patches, and a 64-fold ($8^2$) increase in token comparisons.
This complexity hinders the widespread training and development of Transformer encoders for 3D biomedical images with common hardware, namely CPUs and single GPUs.
The result is a reduced pool of laboratories able to reproduce and build on SOTA models.
In order to encase Transformers blocks into efficient neural networks, we adapt TokenLearner and TokenFuser~\cite{ryoo2021tokenlearner} bottlenecks and insert them between UNet-like, convolutional encoder and decoder.
The last encoder feature map is fed to TokenLearner, a module originally introduced for 2D Vision and Video Transformers (3D input but 1 dimension is time).
TokenLearner classifies each of $p$ pixels (or voxels) as more or less relevant to a set of $N$ abstract classes, with $N\ll p$.
This process yields $N$ spatial attention maps, and $N$ token embeddings are pooled from feature maps according to attention scores.
Token embeddings at this stage can be fed to any token processing architecture, such as Transformers (even pre-trained with self-supervision) and MLP-Mixers~\cite{tolstikhin2021mlp}.
The TokenFuser module brings back token information to 2D or 3D space, for the processing of downstream layers.
Analogous to TokenLearner, TokenFuser generates attention maps by classifying voxels as pertinent to the $N$ tokens' semantic classes.
Then attention masks and tokens are mixed into a new feature map.
Our results confirm how TokenLearner and TokenFuser allow to integrate Transformers into virtually any 3D CNN autoencoder, and how the integration improves simple and efficient UNets, topping performances of slower and memory-heavier SwinUNETR, with easier requirements.
The evidence suggests the effectiveness and viability of TokenLearner and TokenFuser as tokenization methods to reduce memory footprint of training and inference, allowing development on common hardware available to researchers.
The method is naturally interpretable, thanks to its attention maps that encode the location and impact of voxels for the neural network output.
The information bottleneck~\cite{goldfeld2020information} embodied by the tokenization and detokenization may~\cite{saxe2019information} nudge the network towards better representations, which leaves the possibility of better adaptation of Transformers in this framework.

\section{Methods}
This section describes the architecture choice and the training setting of our experiments.
We first develop an effective UNet variant, identified as UNet**, based on the observation that downsampling-upsampling is necessary for speed of computation, decrease of memory footprint, and performance metrics, while concatenating skip-connections can be switched to additive skip-connections with no loss of accuracy and relatively decreasing memory usage.
We then evaluate TokenLearner and TokenFuser as bottlenecks. 
Incorporating them constitutes the Token-UNet, specifically without Transformer.
Finally, we encase a small Transformer encoder between the two Token modules, for the Token-UNet with Transformer variant.
It is important to note that other token processing layers may be included for new Token-UNet variants. 
Our 3 architectures are compared to a vanilla UNet and a SwinUNETR implemented according to published settings, on a 5-fold Cross Validation with 60 epochs per fold (sufficient to stabilize the loss values).
In accordance with seeking low computational costs, hyperparameter tuning and search are avoided.


\subsection{Architectures and modules}

\subsubsection{UNet}
The UNet architecture, introduced in \cite{ronneberger2015unet}, is originally a 2D CNN inspired by Fully Convolutional Neural Networks, with a contracting path or encoder, and an expansive path, or decoder.
Each encoder block is constituted by a convolutional layer with $3\times3$ spatial kernels and doubling channel length, a nonlinear activation (ReLU), and a max pooling for downsampling feature maps.
Each decoder block is instead comprised by a transposed convolution with $2\times2$ spatial kernels and halving channel length, the skip-connection concatenating the encoder feature map of the same spatial resolution, followed by two rounds of convolutional layer with $3\times3$ spatial kernels and nonlinear activation.
UNet-like architectures in general maintain the original distinctions in contracting path and expansive path, however downsampling, upsampling, skip-connections and configurations of blocks may vary from instance to instance and integrate several modifications from innovative architectures in computer vision.
In this work, the default UNet architecture is considered the configuration arising from the MONAI framework integration of the work presented in \cite{isensee2021nnu}, that adds instance normalization layers, LeakyReLU activations, and maintains a constant $3\times3\times3$ size for all 3D convolutional kernels, in both paths.
The strengths of these configurations should be highlighted.
First, downsampling then upsampling along the spatial dimensions allows to reduce the memory footprint and the time required for each convolutional layer to process its input.
Encoder layers that double feature size and halve resolution over 3 spatial dimensions result in 4-fold reduction of tensor ``volumes''.
Second, the larger number of layers and parameters allowed makes the networks more expressive, at the theoretical cost of more compute.
It must be noted a lack optimized CUDA kernels calls from PyTorch, for operations with less parameters such as grouped, depthwise and spatially separable convolutions.
The regularizing effect of less parameters could increase generalization performance, with lower computational cost.
However in practice non optimized operations require more compute time, without compressing notably the memory footprint, while underfitting.
Thus, we do not target convolutional operations to improve the UNet blueprint.
Our proposed UNet-like case modifies the original design in the basic building block, as well as in the connections between blocks.
The building block is a residual function \cite{he2016deep} composed by $3\times3\times3$ convolution, instance normalization, nonlinear activation (GELU \cite{hendrycks2016gaussian}).
The residual is added to an identity mapping of the input, when the output has the same dimensions; when spatial downsampling is needed, average pooling is performed, while trilinear interpolation is employed for spatial upsampling; in case of changes in channel dimensions, pointwise ($1\times1\times1$) convolution is applied to extend or compress the original number of channels.
This ensures a clean path for backpropagated gradients and the possibility of preserving most of the input information where it may be needed across the forward path.
Moreover, all skip-connections from encoder to decoder are additive, exactly as those in-block, instead of being concatenating.
This change allows for approximately halving the memory footprint, speed, and parameter count of the expansive path, with no loss of expressiveness and little overhead.

\subsubsection{TokenLearner and TokenFuser}
The Token-UNet variants encase a TokenLearner and TokenFuser module between the CNN encoder and the MLP classifier, as shown in \ref{fig:tokenarch}.
TokenLearner is based on a simple idea: distant pixels or patches of an image can share the same features, constitute the same structure, or belong to the same abstract class.
On this premise, it is possible to select such pixels and pool the original image to aggregate only this specific information, regardless of distance and dismissing irrelevant noise in the vicinity.
Instead using 1 token per patch of neighboring pixels (voxels), 1 token can represent a size-independent set of variously akin pixels.
In practice, TokenLearner employs a Multi-Layer Perceptron (MLP) to evaluate each image element's pertinence to $N$ non exclusive classes, based on its vector of features.
The soft classification into $N$ categories yields $N$ spatial attention masks: each pixel (voxel, patch) location has an attention score, serving as a class logit, but normalized over space.
With 3D image feature map $X$ of height $H$, width $W$, depth $D$, nonlinear projection MLP$_{\text{TL}}$, and spatial attention masks collected in $A_N$:
\begin{align}
    X \in \mathbb{R}^{F\times H\times W\times D} \nonumber\\
    \text{MLP}_{\text{TL}}: \mathbb{R}^{F...} \mapsto \mathbb{R}^{N...} \nonumber\\
    A_N = \underset{H,W,D}{\text{softmax}}\left(\text{MLP}_{\text{TL}}(X)\right),  \quad \text{with } A_N \in \mathbb{R}^{N\times H\times W\times D}. 
\end{align} 
The $N$ sets of softmaxed attention scores act as weights for $N$ global average poolings of the image tensor, yielding $N$ global vectors, each focusing on a semantic group of pixels.
Using a simplified Einstein notation:
\begin{equation}
    T = A_N X, \quad t_{nf} = a_{nhwd}x_{fhwd}, \quad \text{with } T \in \mathbb{R}^{N\times F}.
\end{equation}
Equivalently, one can see each spatial attention mask $A_n$ with $n\in \{ 1,2,...,N\}$ broadcast over all $F$ feature channels with operator $\beta_F(\cdot)$, followed by sum reduction of the spatial dimensions, thus $t_n=\beta_F(A_n)X$. 

The TokenFuser module is introduced to transform the $N$ tokens back into the original shape and resolution of the input.
One could perform "unpooling" of the tokens, by an outer product between tokens and original attention masks, effectively tiling the token for each $n$-th class  of $N$ into the original tensor shape, and then re-weighting according to the original contribution of a pixel for class $n$.
Instead of this parameter-free unpooling, in TokenFuser a new MLP outputs new spatial attention masks.
Crucial for the effectiveness of TokenFuser, a mixing matrix $M$ linearly combines global tokens before unpooling.
\begin{align}
    \text{MLP}_{\text{TF}}: \mathbb{R}^{F...} \mapsto \mathbb{R}^{N...} \nonumber\\
    B_N = \text{MLP}_{\text{TF}}(X),  \quad \text{with } B_N \in \mathbb{R}^{N\times H\times W\times D} \nonumber\\
    Y = B_N MT, \quad y_{fhwd} = b_{nhwd}m_{nn}t_{nf}, \quad Y \in \mathbb{R}^{F\times H\times W\times D} \nonumber\\
    X \leftarrow X + Y 
\end{align}
In this work $N:=8$, as in the original work, as the semantically relevant elements for segmentation are at least 3 (the number of ground truth labels).

\subsubsection{Transformers}
The tokens extracted by TokenLearner are processed by a typical Transformer architecture with $4$ Transformer encoder blocks, each made of 2 residual blocks: MultiHead Self-Attention (MHA), and MLP as Feed-ForwardF Network (FFN).
The model size determines the width of matrices in the MHA and FFN modules in the Transformer blocks, repeating the following structure for every $i$-th block:
\begin{align}
    X_{\text{MHA}, i} \leftarrow X_{\text{FFN}, i-1} + \text{MHA}(\text{LN}(X_{\text{FFN}, i-1})) \nonumber\\ 
    X_{\text{FFN}, i} \leftarrow X_{\text{MHA}, i} + \text{FFN}(\text{LN}(X_{\text{MHA}, i})),
\end{align}
with 
\begin{align}
    \text{FFN}: \text{MLP}(X)=(\phi(XW_1))W_2 \nonumber\\
    \text{MHA}: \left(\underset{h\in H}{\text{concat}} \left[ \text{SA}_h \right]  \right)W_O \nonumber\\
    \text{SA}: \text{softmax}\left(\frac{XW_{Q}W_{K}^TX^T}{\sqrt{d/h}} \right)W_{V} \nonumber\\
    \text{LN}: \gamma \left(\frac{X - \mu_X}{\sigma_X} \right) + \beta.
\end{align}
In the present model, for every MLP, $W_1 \in \mathbb{R}^{d\times 4d},W_2 \in \mathbb{R}^{4d\times d}$, and $\phi(\cdot)$ is the GELU nonlinearity~\cite{hendrycks2016gaussian}; for every SA head and MHA module respectively, $W_Q, W_K, W_V \in \mathbb{R}^{d/h\times d/h}$ and $W_O \in \mathbb{R}^{d\times d}$, with $d$ model dimension (the number of features for tokens, thus $d=F=256$ of the convolutional encoder) and $h=8$ number of heads.

\begin{figure}[!ht]
    \centering
    \includegraphics[width=0.9\linewidth]{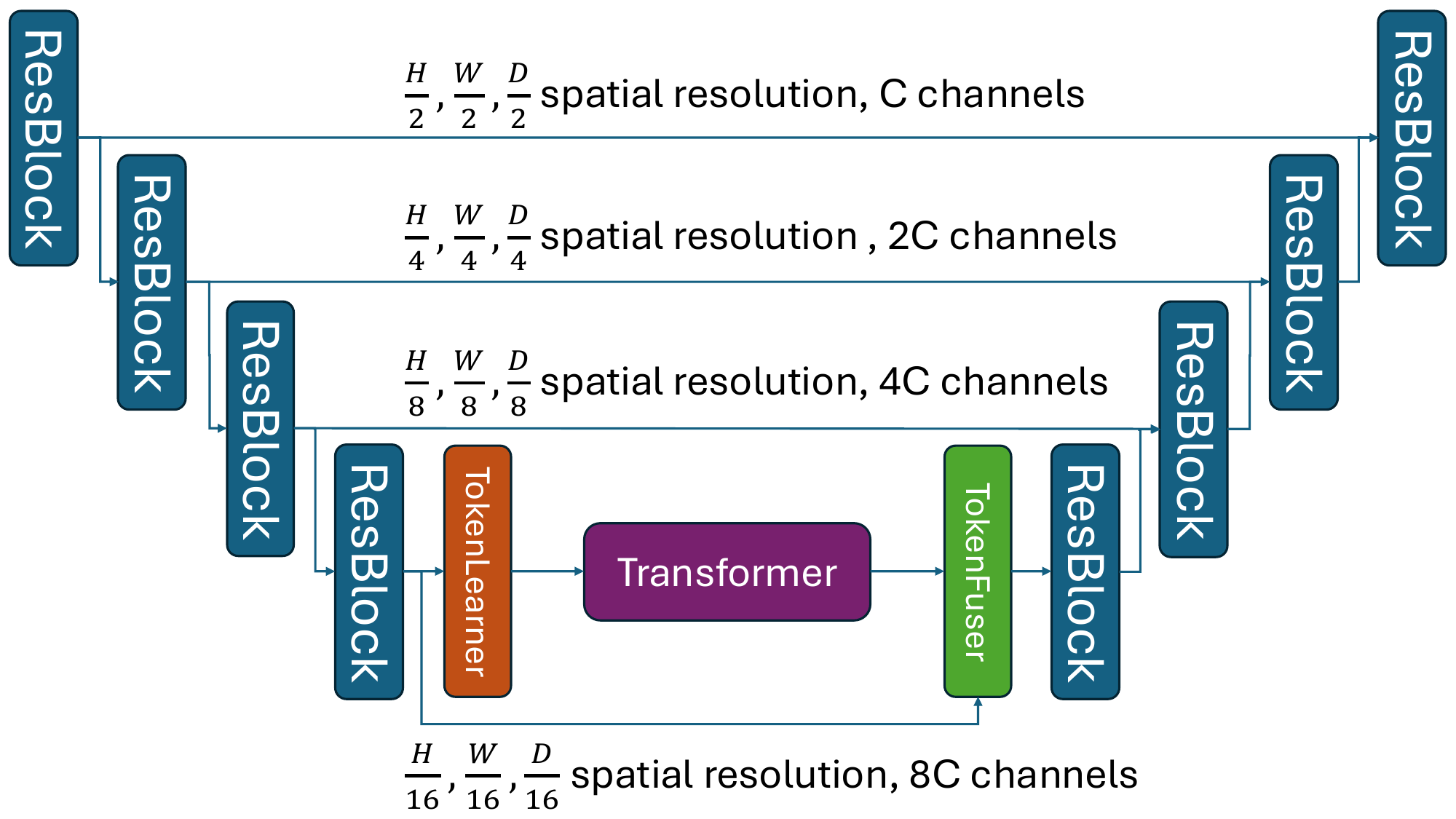}
    \caption{Architecture of Token-UNet including the encased Transformer.}
    \label{fig:tokenarch}
\end{figure}

\begin{figure}[!ht]
    \centering
    \includegraphics[width=0.9\linewidth]{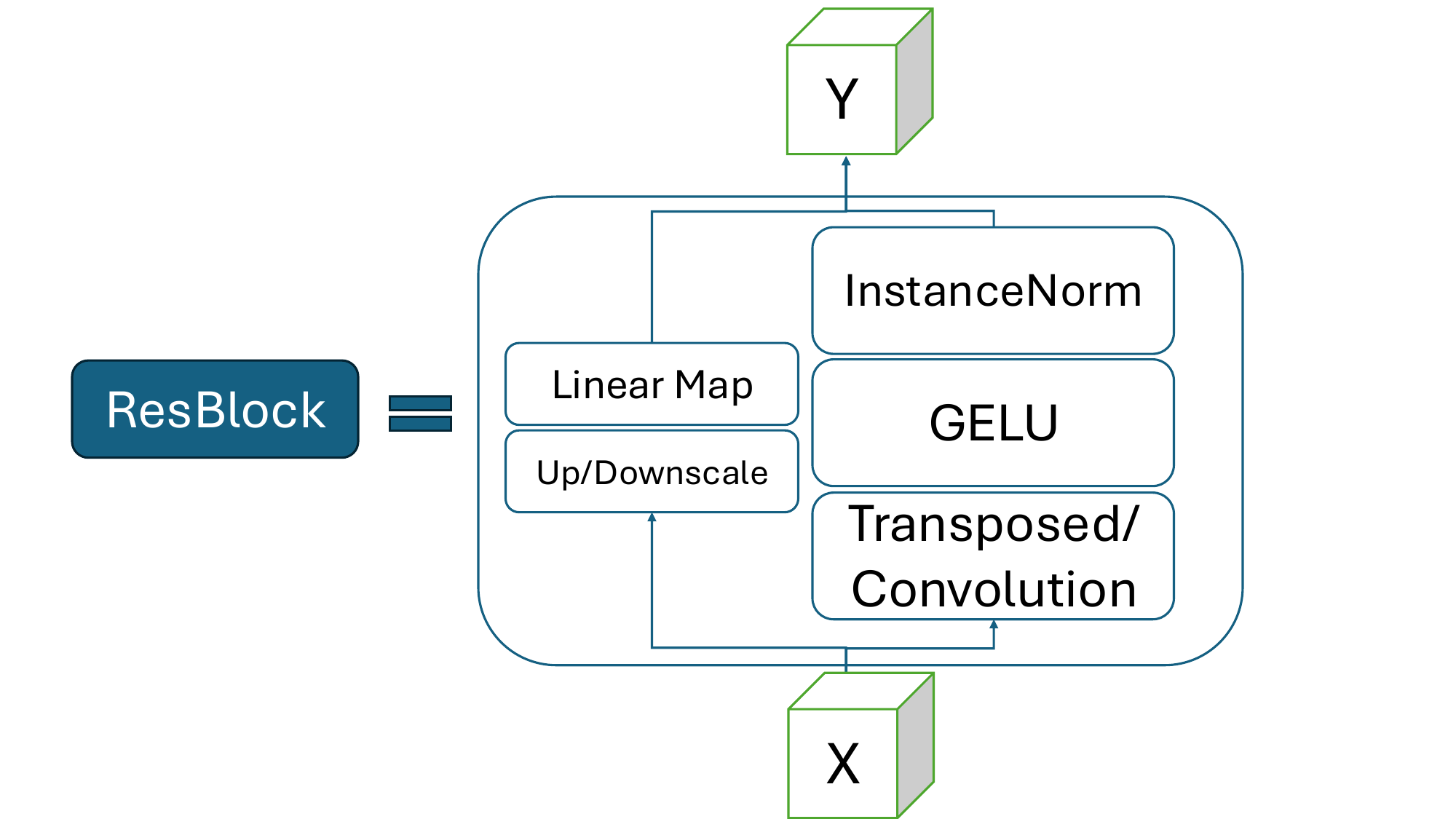}
    \caption{General form of our ResBlocks, with options to upsample or downsample feature maps for the residual paths (with strided convolutions or transposed convolutions) and internal skip-connections (with average pooling or linear upsampling.}
    \label{fig:resblock}
\end{figure}

\begin{figure}[!ht]
    \centering
    \begin{subfigure}{0.49\textwidth}
        \includegraphics[width=0.98\linewidth]{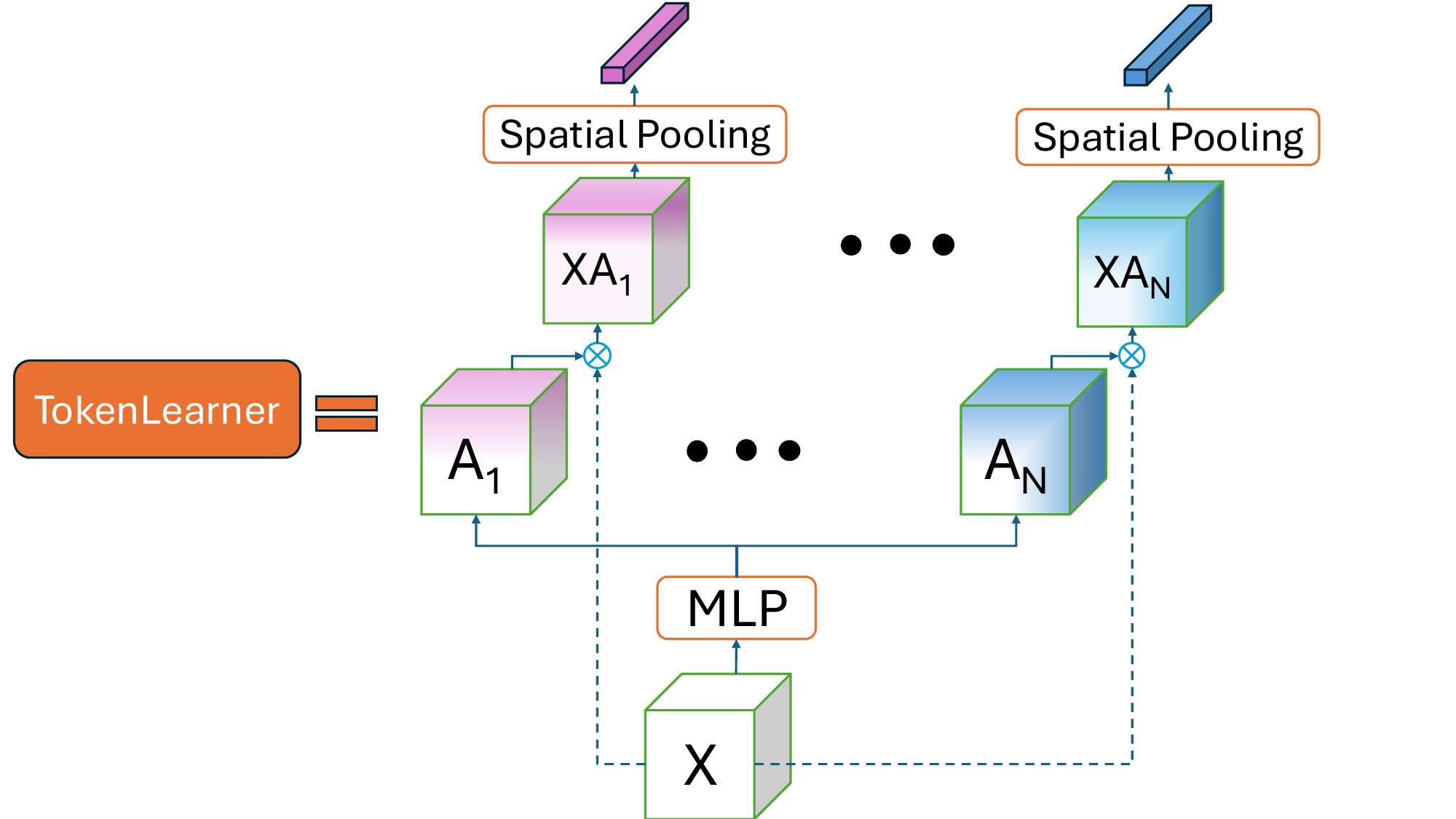}
        \caption{}
    \end{subfigure}
    \begin{subfigure}{0.49\textwidth}
        \includegraphics[width=0.98\linewidth]{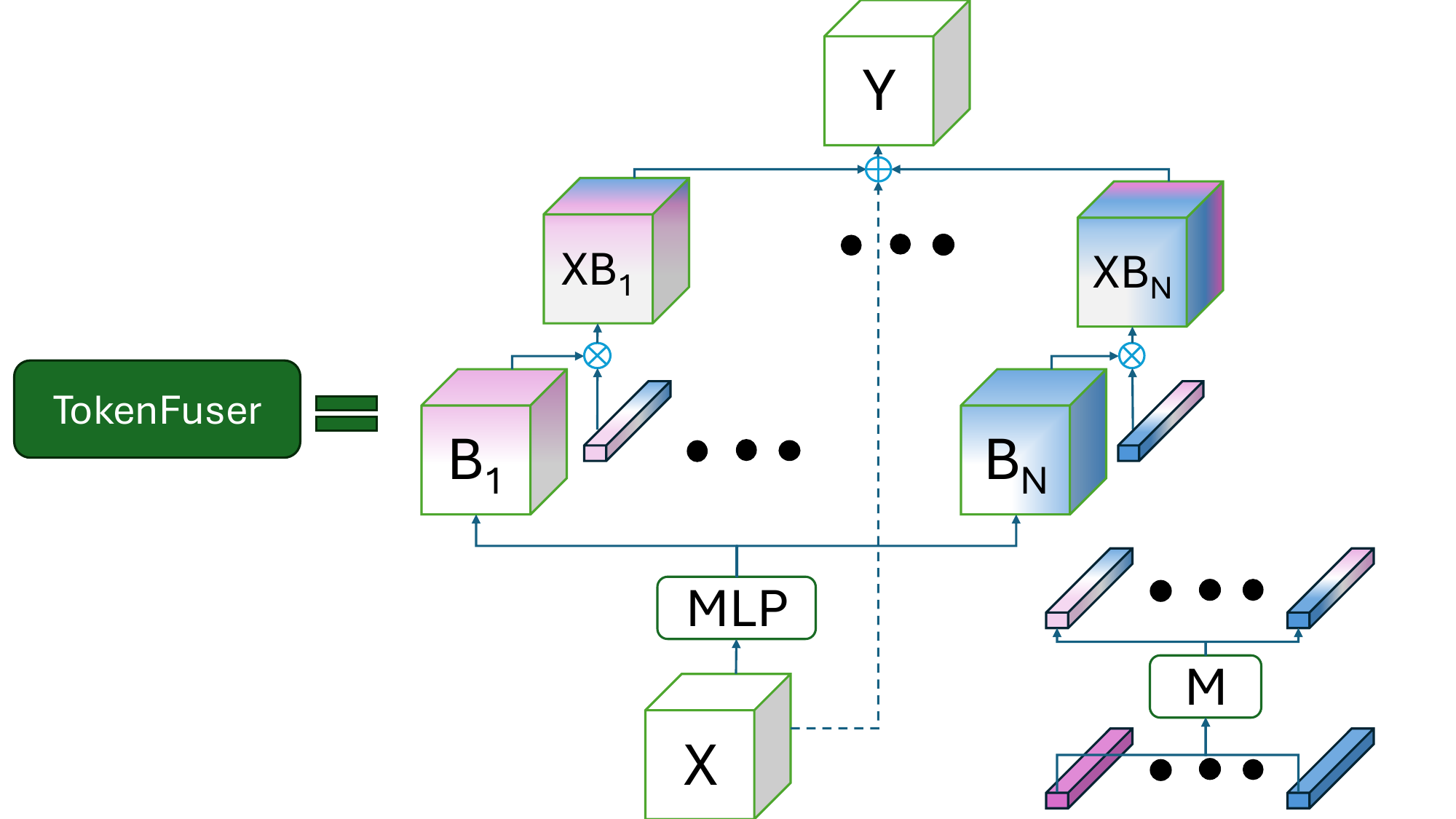}
        \caption{}
    \end{subfigure}
    \caption{The last encoded cube is fed to TokenLearner and transformed into N=8 different pooled vectors that encode semantic information from all pertinent locations. These vectors eventually serve as tokens in the Transformer. TokenFuser than recreates N=8 spatial masks for tokens, that are linearly mixed, then broadcast over the masks and finally summed to the last encoded cube, that will be decoded by the ascending UNet path.}
    \label{fig:tokenlearnfuse}
\end{figure}

Figure \ref{fig:tokenarch} shows the overall architecture of a Transformer in a Token-UNet, while Figure \ref{fig:resblock} and \ref{fig:tokenlearnfuse} illustrate the building blocks and tensors transformations.


Table \ref{tab1:pms} summarizes the different variants and sizes, with their respective parameter counts (M for $10^6$ parameters).
\begin{table}[!ht]
\centering
\begin{tabular}{lccccc}
\toprule
\textbf{UNet} & \textbf{SwinUNETR} & \textbf{UNet**} & \textbf{Token-UNet (w/o Transformer)} & \textbf{Token-UNet (w/ Transformer)} \\
\midrule
 12.89M & 15.71M & 2.42M & 2.45M & 5.51M \\
\bottomrule
\end{tabular}
\caption{Trainable parameter counts.}\label{tab1:pms}
\end{table}

\begin{table}[!ht]
\centering
\footnotesize
\begin{tabular}{lccc}
\toprule
 & \textbf{new-UNet} & \textbf{Token-UNet (TL-TF only)} & \textbf{Token-UNet} \\
\midrule
\textbf{Tokenizer}   &   &  \checkmark &  \checkmark \\
\textbf{Transformer}   &   &   &  \checkmark \\
\textbf{Detokenizer}   &   &  \checkmark &  \checkmark \\
\textbf{Pre-trained weights}   &   &   &   \\
\bottomrule
\end{tabular}
\caption{Components and incremental additions from UNet to Token-UNet.}\label{tab2:modules}
\end{table}
\normalsize

\subsection{Data}
The dataset is the FeTS 2022 Challenge Dataset, a subset of BraTS Continuous Evaluation Challenge Dataset, comprising 1251 subjects with radiographically appearing glioblastoma.
Brain scans are collected in routine clinical acquisitions from several institutions, pre-operation, with multi-parametric Magnetic Resonance Imaging (mpMRI).
For each subject, 4 modalities are available, from the MRI sequences: T1-weighted, T1-weighted Gadolinium post-contrast enhanced, T2-weighted, and T2-FLAIR.
The original sequences vary in slice thickness and axial acquisition, but all images are resized and resampled to $240\times240\times155$ resolution, 1mm$^3$ voxel size.
Ground truth annotations of the tumor sub-regions are from expert neuroradiologists.
Tumor labels are WT (whole tumor), TC (tumor core), and AT (active tumor).
Denominations refer to tissue properties and appearance in the available modalities.
Further information can be found at the Challenges sites\footnote{https://www.synapse.org/Synapse:syn28546456/wiki/617093}\footnote{https://www.synapse.org/Synapse:syn27046444/wiki/616571} and in~\cite{Bakas2018IdentifyingTB}.

\subsection{Training}
We train and test our models with a 5-fold Cross Validation over the BraTS training set, since the Challenge validation data are unlabeled.
The subjects are randomly divided into 5 folds, and this division is maintained across experiments with different model sizes and weight initialization schemes.
Each fold training runs for 60 epochs.
The learning objective minimizes an averaged Dice and cross-entropy loss.
The UNet and Token-UNet family models are optimized with SGD with Nesterov momentum \cite{nesterov1983method}~\cite{nesterov2013introductory}, with learning rate set to $10^{-2}$, while the SwinUNETR models are optimized with AdamW ~\cite{loshchilov2017fixing} and learning rate of $8\cdot10^{-4}$.
All learning rates are reduced with a cosine annealing schedule.
The actual batch size is set to 1 to alleviate the memory footprint, with gradient accumulation for 16 mini-batches to increase the effective batch size with no loss of generality, since none of the models employ layers based on batch statistics.
During training, patches of size $128\times128\times 128$, randomly extracted from the images, are fed to the model, once per subject; during evaluation, sliding window inference on $128\times128\times 128$ windows is integrated over the whole 3D brain scans.
The final performance is measured in terms of Dice score, only.
Experiments are implemented in PyTorch\footnote{\url{https://pytorch.org/}} (version 2.0.1) and MONAI\footnote{\url{https://monai.io}} (version 1.2.0).
Experiments are run on a computing node with 64 Intel(R) Xeon(R) Silver 4314 CPU @ 2.40GHz, connected to a NVIDIA A30 Tensor Core GPU with 24 GB of GPU memory.
The code to replicate our results will be made available at [...] upon publication.

\section{Results}
The results in this section resume the parameter and memory footprint of Token-UNets compared to UNet, UNet** and SwinUNETR, as well as the average Dice score performance of model variants, and the visualization of TokenLearner and TokenFuser outputs.

\subsection{Comparing parameter size and memory with (Swin)UNETR}
We implement UNet and SwinUNETR according to the respective papers training settings and common library defaults, using the MONAI framework.

Regarding UNet, setting a convolutional feature size of $32$ at maximum resolution and $320$ at minimum resolution yields almost $13$ million parameters (Mpms).
The space in memory during training is $1.2$GB with batch size of 1. 

Regarding SwinUNETR, setting a convolutional feature size entails the same token feature size, which starts at $24$, i.e. smaller than Swin-T, the tiny version of 2D SwinTransformers~\cite{liu2021swin}.
The SwinTransformer employs a mechanism to reduce the number of tokens, by concatenating neighboring tokens along the feature axis, which is then linearly projected to double the previous layer feature size.
The encoder thus has 4 merge-and-double bottlenecks, and the feature size is $24, 48, 96, 192$.
Moreover, SwinUNETR is trained on $128\times 128\times 128$ resolution, affecting the number of positional encodings (relative position biases) required by the SwinTransformer.
This SwinUNETR reaches $15$ Mpms, and training requires $14$GB with batch size of 1.

For what concerns the Token-UNet family, the blueprint case is our additive UNet**, where we remove the most internal stages of feature size $320$ for a two-fold reason: on the one hand, it is preferable to tokenize a relatively high-resolution feature size, with rich spatial information; on the other hand, the feature size of $256$ is more common both in convolutional and Transformer models, making it easier to integrate pretrained models, eventually.
The comparison is at worst favorable towards the original UNet rather than our additive UNet, and yet the latter performs better even without tokenization, despite having less parameters.
In our cross-validated models, GPU memory is occupied mostly by convolutional features that are set to token feature size by default, without further decoupling techniques.
It is possible to instantiate a "decoupled" Token-UNet, setting convolutional feature size ${d_c}$ and expanding tokens linearly to a token feature size ${d_t}$ to possibly match several pretrained Transformers or token-mixing architectures of any size.
An extending linear projection $W_e(T): \mathbb{R}^{d_c}\mapsto \mathbb{R}^{d_t}$ can be added after TokenLearner, and a shrinking linear projection $W_s(T): \mathbb{R}^{d_t}\mapsto \mathbb{R}^{d_c}$ added before TokenFuser.

\begin{figure}[!ht]
    \centering
    \includegraphics[width=0.8\linewidth]{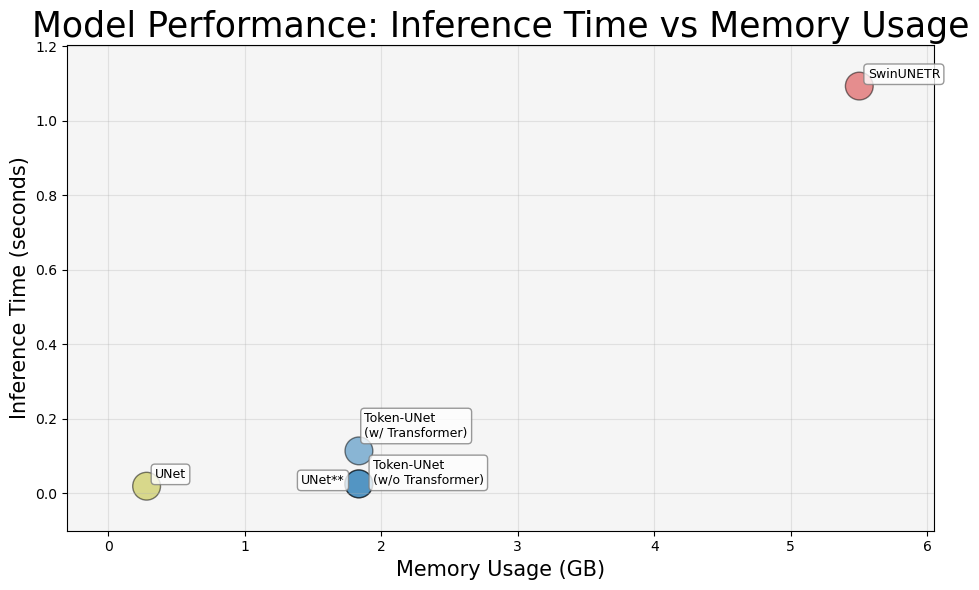}
    \caption{Inference time vs inference memory occupation on GPU.}
    \label{fig:tradeoffs}
\end{figure}

Figure \ref{fig:tradeoffs} shows how the UNet has the fastest and lightest inference, while SwinUNETR has the slowest and heaviest.
UNet** and Token-UNet without Transformer overlap, showing how TokenLearner and TokenFuser are almost free of cost at inference time.
Token-UNet models are very close to one another, showing how light a Transformer can be when the number of tokens is constrained.
These models reach and top SwinUNETR in terms of accuracy, while being not far from UNet in hardware and time requirements.

\subsection{Convergence}
\begin{figure}
    \centering
    \begin{subfigure}{0.9\textwidth}
        \centering
        \includegraphics[width=0.8\linewidth]{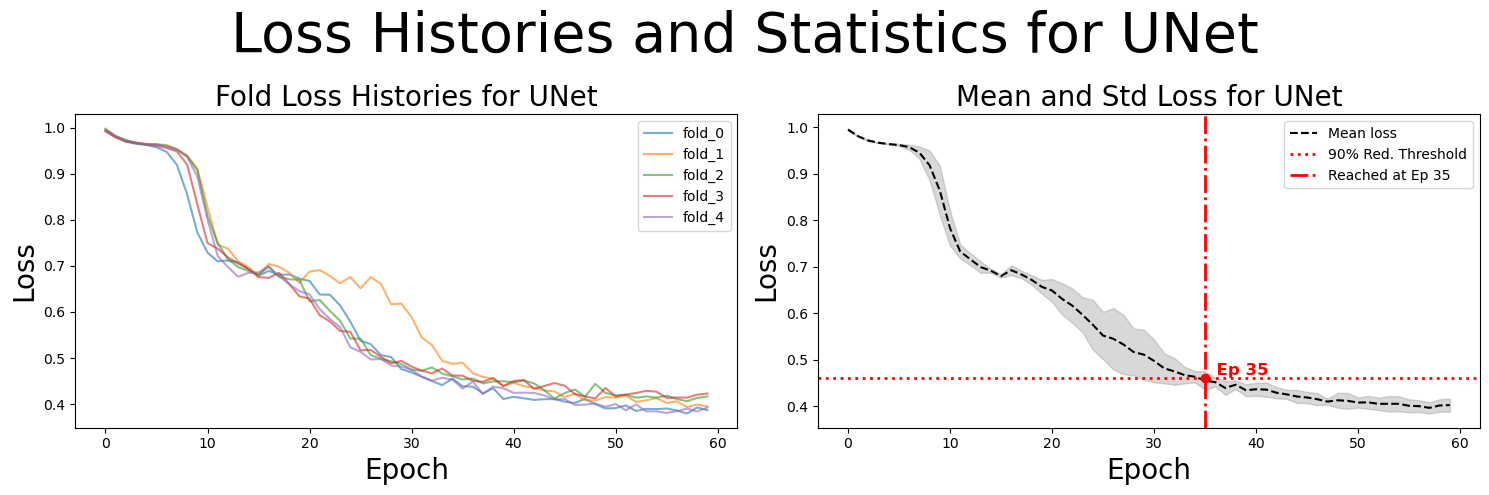}
    \end{subfigure}
    \begin{subfigure}{0.9\textwidth}
        \centering
        \includegraphics[width=0.8\linewidth]{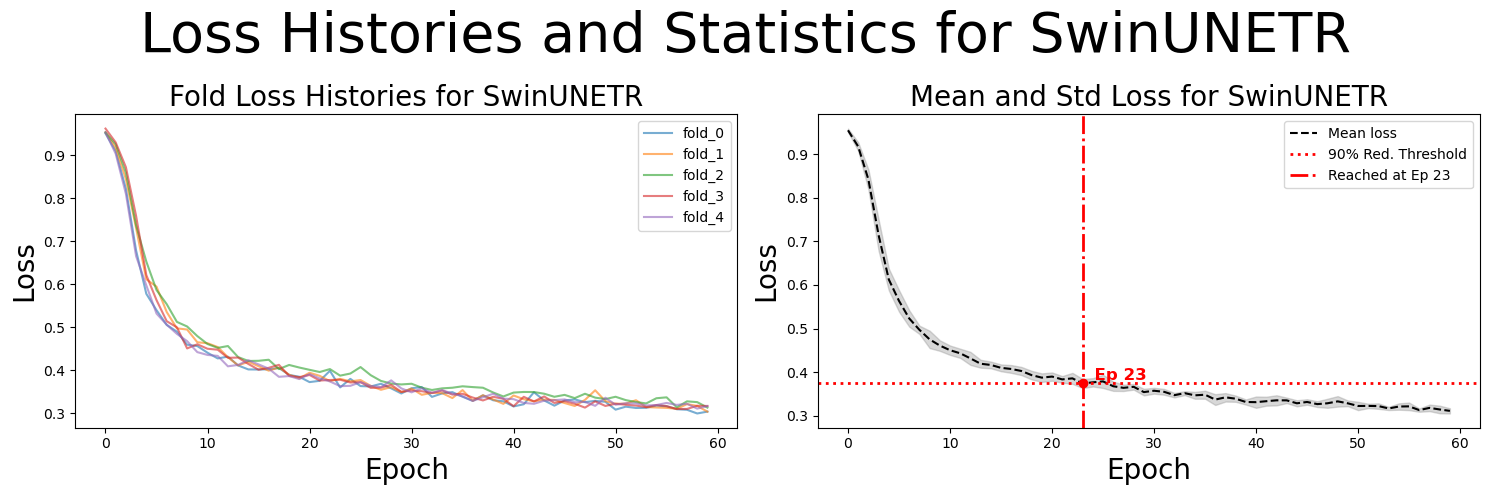}
    \end{subfigure}
    \begin{subfigure}{0.9\textwidth}
        \centering
        \includegraphics[width=0.8\linewidth]{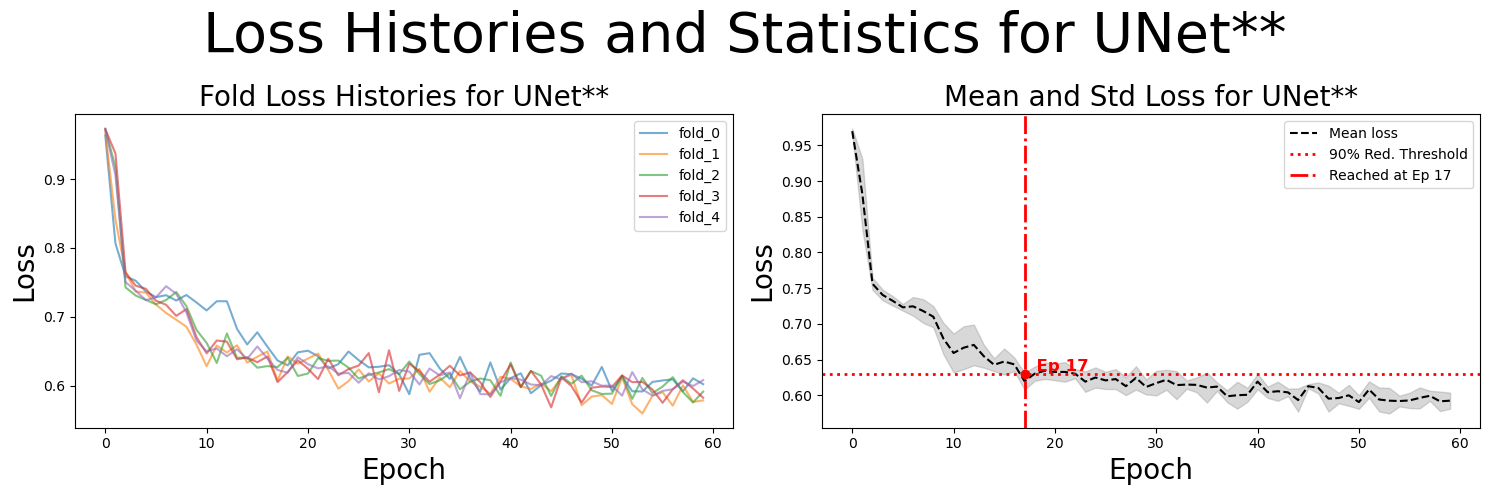}
    \end{subfigure}
    \begin{subfigure}{0.9\textwidth}
        \centering
        \includegraphics[width=0.92\linewidth]{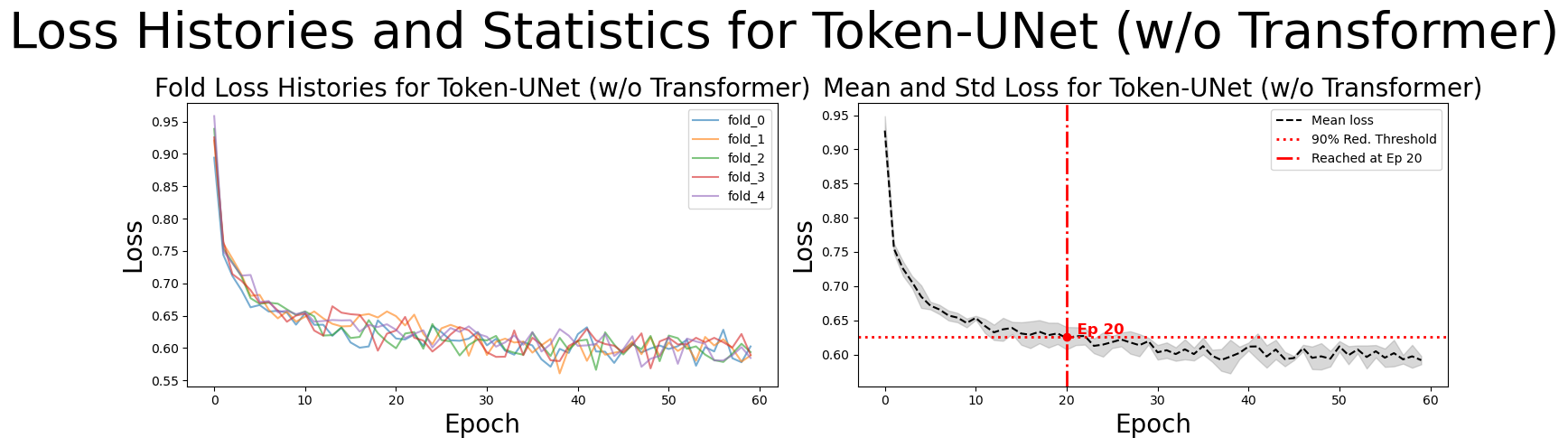}
    \end{subfigure}
    \begin{subfigure}{0.9\textwidth}
        \centering
        \includegraphics[width=0.92\linewidth]{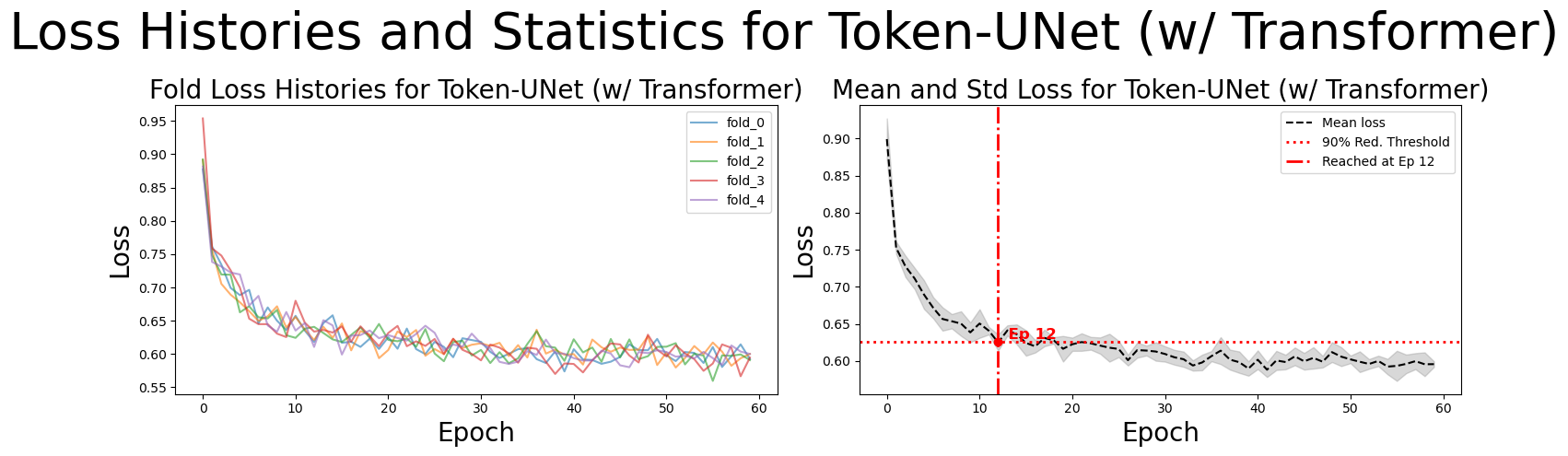}
    \end{subfigure}
    
    \caption{Loss history by fold for every architecture.}
    \label{fig:losses}
\end{figure}

\begin{figure}[!ht]
    \centering
    \includegraphics[width=0.8\linewidth]{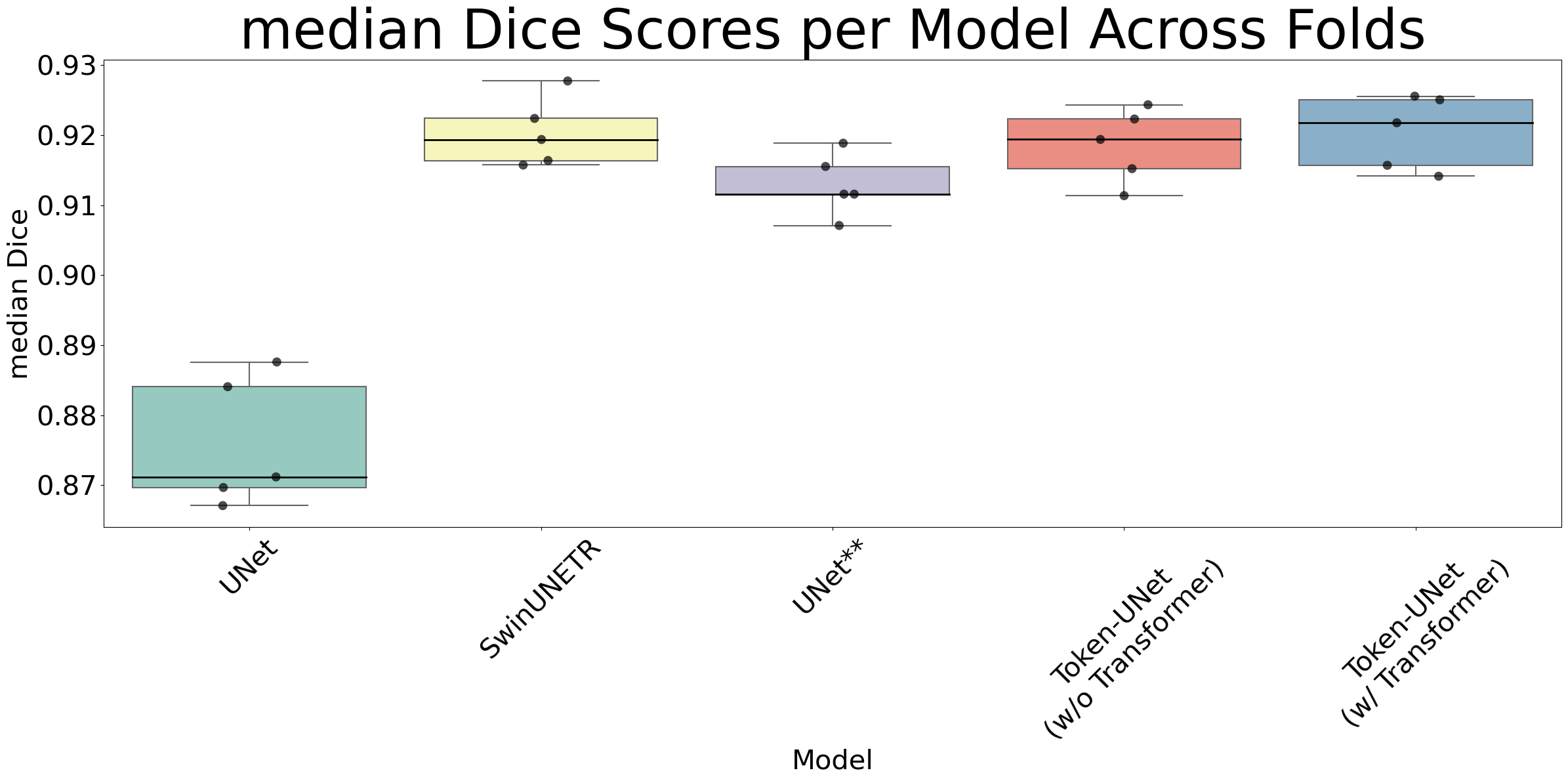}
    \caption{Boxplots of Dice scores obtained by each model during cross-validation. The changes from classic UNet to our UNet** make the single most relevant performance increase. In the Token-UNet family the addition of TokenLearner and TokenFuser is more impactful than the addition of a Transformer.}
    \label{fig:cv}
\end{figure}

The cross-validated models show mostly typical loss history patterns, with fast convergence and diminishing returns in loss reduction with more epochs.
Figure \ref{fig:losses} shows each architecture's loss patterns.
Our models converge faster towards their final loss value, as exemplified by the epoch at which the average loss across folds goes below than 90\% of total loss reduction between initial and final loss, which can be viewed as the speed to reach 90\% of final performance.
Despite final loss values being higher for our models, performances reach and even surpass those of SwinUNETR and UNet on validation data, as seen in Figure \ref{fig:cv}.

Figure \ref{fig:cv} shows the cross-validation boxplot for the 20 models trained and evaluated (4 architecture variants $\times$ 5 validation folds).
Larger models do not necessarily have the better median performance.
Accounting for parameter efficiency, the stand-alone addition of TokenLearner and TokenFuser, seen in the Token-UNet models, is the most effective enhancement of models.
In this setting, TokenLearner acts as an information bottleneck that appears to increase the semantic throughput of models.
\subsection{What TokenLearner learns}
\begin{figure}[!ht]
    \centering
    \begin{minipage}{0.5\textwidth}
        \centering
        \includegraphics[width=\linewidth]{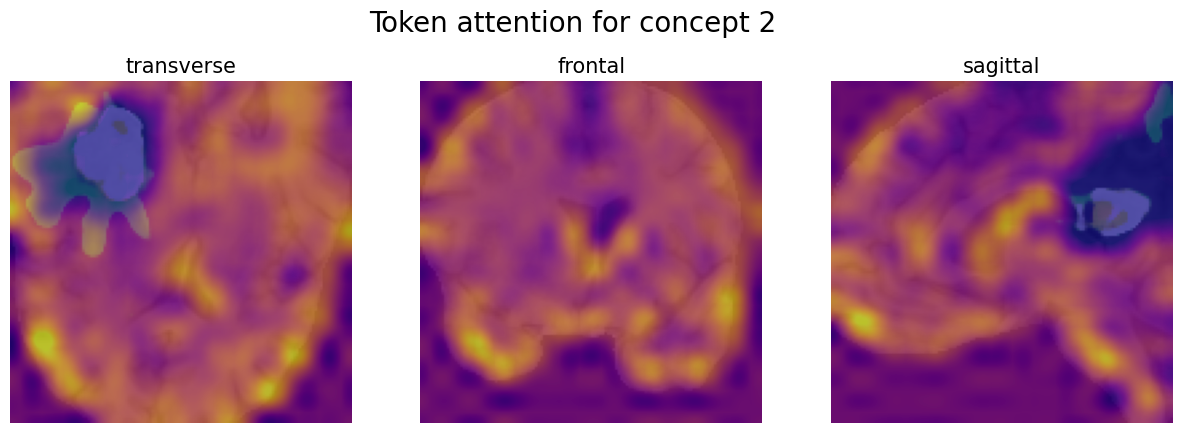}
    \end{minipage}%
    \begin{minipage}{0.5\textwidth}
        \centering
        \includegraphics[width=\linewidth]{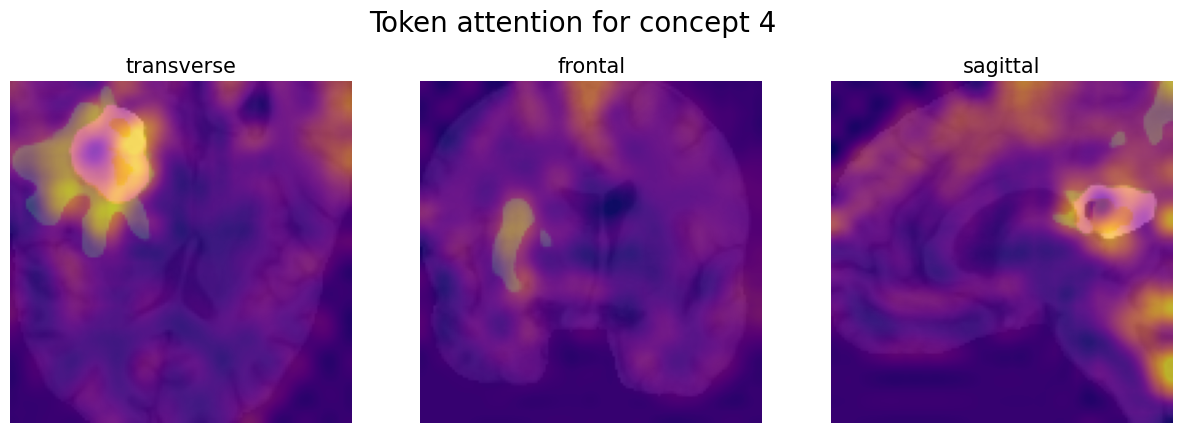}
    \end{minipage}

    \begin{minipage}{0.5\textwidth}
        \centering
        \includegraphics[width=\linewidth]{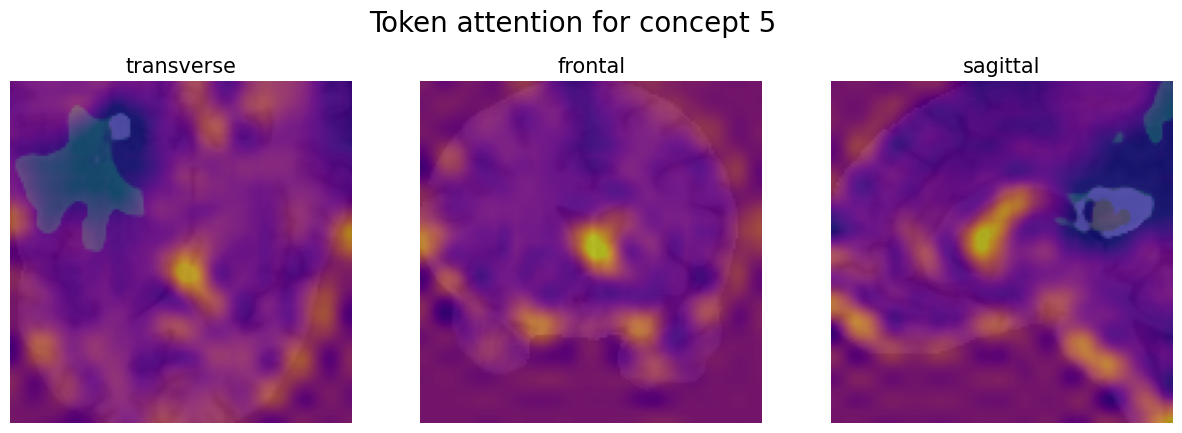}
    \end{minipage}%
    \begin{minipage}{0.5\textwidth}
        \centering
        \includegraphics[width=\linewidth]{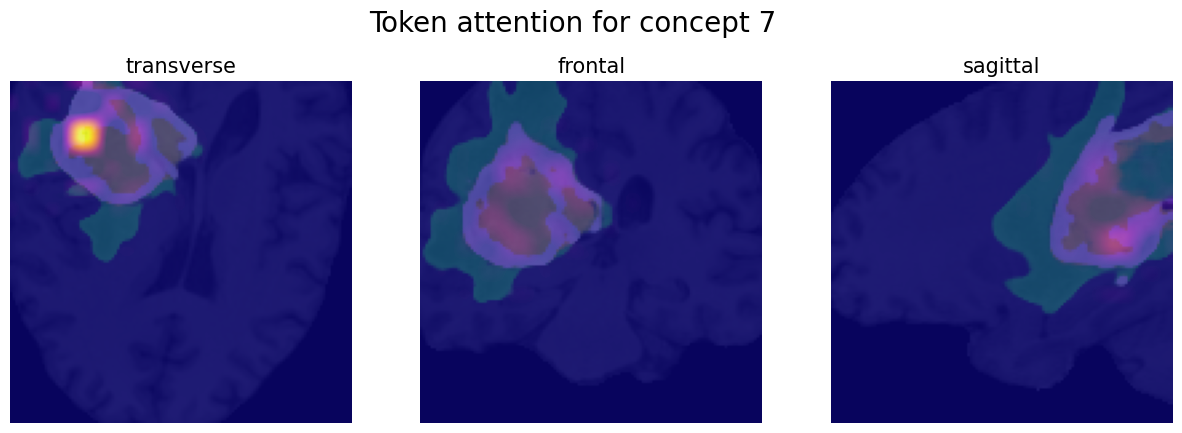}
    \end{minipage}

    \begin{minipage}{0.5\textwidth}
        \centering
        \includegraphics[width=\linewidth]{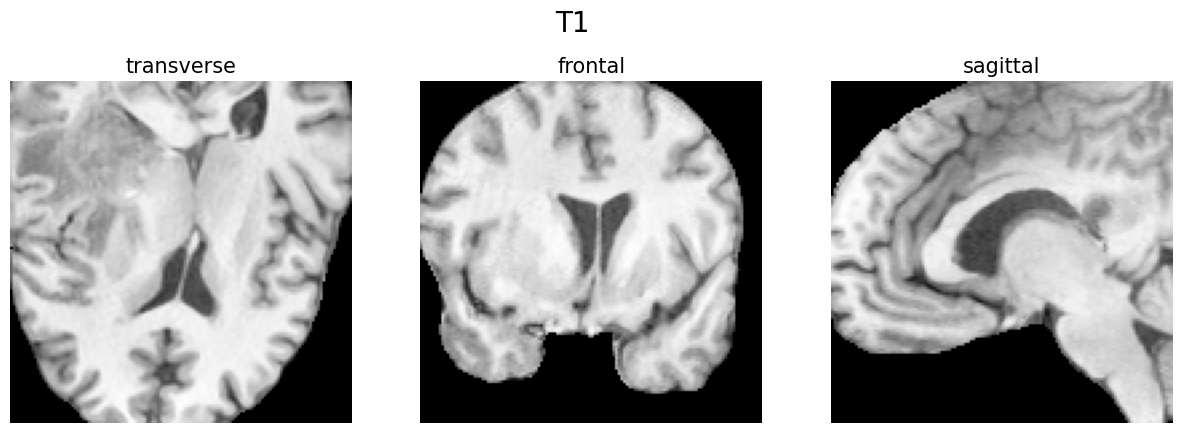}
    \end{minipage}%
    \begin{minipage}{0.5\textwidth}
        \centering
        \includegraphics[width=\linewidth]{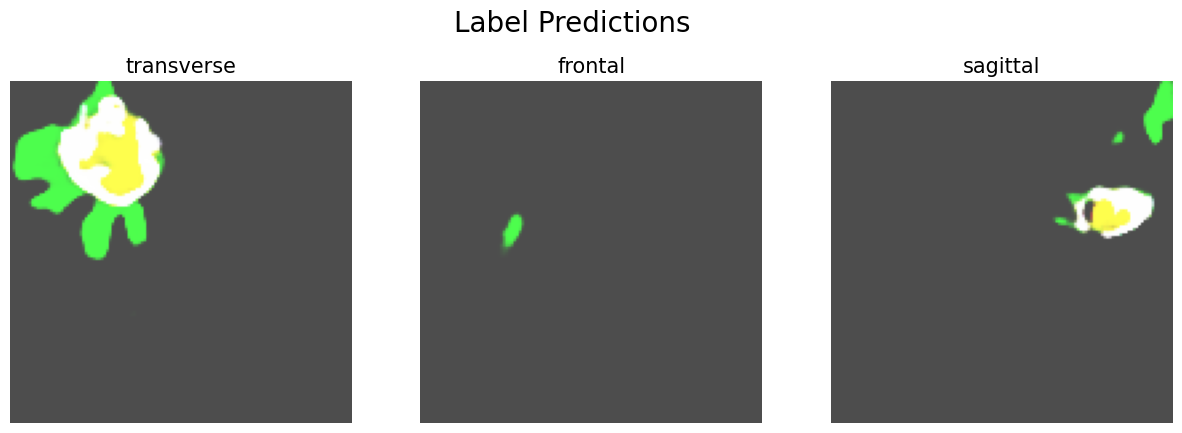}
    \end{minipage}

    \caption{First 2 rows: the TokenLearner spatial attention masks generated from the Token-UNet model evaluated on a random subject from the first validation fold. Last row: the T1 scan on the left, the ground truth mask on the right. Most attention masks are correlated with different aspects of the final prediction, other focus on general features, such as the background and contours.}\label{fig2:attnmaps}
\end{figure}

Figure \ref{fig2:attnmaps} illustrates some of the $N=8$ spatial attention maps in tensor $A_N$, and how they relate to the prediction and the ground truth segmentations.
The highest attention scores (in yellow) correspond to the heighest weights when pooling a semantic token from the feature map $X$.
Some tokens receive contributions from the larger lesion area, others focus on specific core tumor tissues, while other attend to edges between high and low intensity of the image, corresponding to ventricules, brain contours against the background.

\section{Discussion}
Our results generally show how Token-UNet, i.e. an additive UNet with TokenLearner and TokenFuser encasing a Transformer, improves performance over vanilla UNet, reaching that of state-of-the-art architectures such as SwinUNETR with parameters in the order of millions instead of tenths of millions, reduced training and inference memory footprint and computing time.
The performance of Token-UNet demonstrates the positive effect of tokenization, emerging even without token mixing architectures such as Transformers.
The first explicit aim of TokenLearner is to reduce and fix a priori the number of tokens to be processed, and thus the memory footprint and computation requirements of downstream blocks, in particular the attention calculations.
This brings high flexibility, since a variable number of voxels can be associated to any specific token, and a variable volume size can be tied to a set number of tokens, whereas ViTs employ fixed associations between the number of pixels (voxels, patches) and tokens.
As a consequence in classic architectures, if the input size is unbound, the number of tokens quickly becomes intractable on most common devices, especially with 3D images.

In addition to efficiency, TokenLearner nudges the convolutional encoder layers towards keeping semantically relevant information.
This is evidenced by its spatial attention maps, that focus on specific tissues and structure relevant to both domain and task.
The phenomenon opens the question on whether semantic tokenization might interact more favorably with in-domain self-supervised learning, compared to simpler but more expensive patch-based tokenization, consisting in flattening and linear transformations.
It is possible that learning to tokenize effectively, independently of a downstream task, would constitute a useful representation learning objective for self-supervised training.
It would be especially resourceful if it were to uncover previously unknown patterns in the data, or showed to capture features that are difficult to explicitly define and track in supervised learning.
In some failure cases, despite the autoencoder is not able to segment a part of the tumor region, the TokenLearner attributes to it high attention values.
This hints at the high capacity and task-effectiveness of the encoder, and offers an interpretable key for failure case analysis, which would be paramount in biomedical settings, especially to prevent practical consequences of automated models failing and allow human expert intervention. 
The attentional maps produced by TokenLearner thus offer support for analysis of model training modes (either failing or successful) and inference, as well as highlighting features that can guide deep learning practicioners and biomedical scientists to bridge performances and new findings.

Future directions comprise the development of more efficient, effective, and flexible modules to lead 3D data to Transformer blocks and token processing architectures.
Token-UNet shows how TokenLearner is easily adapted to any dimensionality, since it is processing elements according to their features (channel dimensions), agnostic to their location and spatial organization.
MLPs, Transformers and other models are suited to process vectors sets and can be explored for learning meaningful transformations of semantic vectors extract from medical imaging.
These architectures become especially effective and adaptable after pretraining, that is typically the most compute intensive stage, while being potentially performed without labels and explicit supervision, addressing the label scarcity that affects the biomedical domain.
Hyperparameter tuning and new decoding strategies can be combined to enhance final performance and interpretability.

Overall, learning to tokenize is a promising path that allows to bridge the simplicity of convolutional layers with the scaling laws and the versatility of Attention modules, in affordable computational settings.
The tokenization approach of Token-UNet is focused not only on semantic pertinence and interpretability, but also on fixing computational needs and fitting the hardware constraints that most research groups encounter whenever they approach deep learning modeling.
Aside from industry and large research centers, and specific competitions, most researchers deal with CPU-only or single-GPU machines that make it increasingly harder to respectively try inference, training, and pretraining of state-of-the-art architectures.
Moreover, these adoptions usually require several adaptations to the specific domain and data at hand, further weighting on time and compute resources with attuning loops.
By bringing training and deployment on small machines, more research directions could be explored by the community.
Development of new models can forgo more iterations given a fixed compute or time budget, allowing statistical testing to choose the most general strategies, avoiding decisions based on random fluctuations, especially in a field characterized by stratified levels of heterogeneity.
In parallel, the adoption of innovations and of established architectures of weights can be made easier than before, based on the encasing framework and the code shared with the paper.

\section{Conclusions}
Modern AI has moved towards development of foundation models, i.e. large sized models trained possibly without supervision on immense datasets, and capable to solve new tasks on new data mostly without retraining, or with small scale fine-tuning, or integration with smaller task-specific modules.
This thread of development is based mainly on the versatility of the Transformer architecture, its effective combination with modern hardware resources, and consequently on large scale dataset and long lasting training of such models.
Biomedical applications of deep learning techniques are often problematic by these terms, because clinical and research computational resources are limited in hardware, bounded in time, and require several domain-dependent adaptations.
At the same time, data such as biomedical images are inherently three-dimensional (e.g. MRI, Computed Tomography, etc.) due to the nature of underlying structures of interest, further increasing the computational requirements for processing.
Methods able to bring the generality of foundation models to such specialized domains have the potential to accelerate research.
In this context, testing the possibility of applying Transformers to different settings is of great importance.
Since the block architecture with MultiHead Attention and MLP is not specific to language or vision, recent works have employed Vision Transformers in computer vision architectures, even extending them directly to 3D data.
However, task- and domain-specific layers need to be integrated in Transformers, such as linear embeddings from 3D patches to tokens (and back), in the case of 3D biomedical scans.
Departing from previous approaches that integrate Transformers in long-time favorite architectures for medical segmentation, we have designed a new frame around the classic ViT.
Our models adapt the encoder and decoder components around the Transformer in a novel way, with the explicit aim of reducing memory and computational footprints compared to the direct tokenization of 3D data based on patching.
The effect is a great reduction in the resources needed to train and test these models, decoupled from the input resolution, from the number of parameter.
In particular, the TokenLearner framework leads to the output of tokens that refer to sparse and interpretable locations of an image, sharing common features.
While a token is usually tied to a patch with a specific location, and progressively integrates information from other patch-tokens, with TokenLearner a token may be related to all separate parts of the image containing a specific tissue, or belonging to the same anatomical structure.
The process is completely data-driven, and naturally lends to interpretability, which is extremely important in diagnostic-like tasks.

This work is only a first step towards developing and democratizing foundation models for biomedical imaging.
Further research is needed to simplify the encoder stem, possibly studying modality-agnostic embeddings, moving the most of information processing into the promising Transformer body.
The number of tokens required by these models is not fine-tuned, and it is still possible to improve performance by reducing or increasing how many tokens are pooled, at what scale, and so forth.
Although TokenLearner naturally yields semantically meaningful tokens, the $N$ categories seem to be restricted not only in number per image, but also in number and type per dataset.
It is possible that the model may encode several groups of $N$ concepts, depending on the instance processed.
However, the simplicity of the MLP in TokenLearner and yet its black-box aspect make it difficult to predict or assess such properties.
Considering these reasons reinforces the importance of widespread experiments, which are allowed by the very nature of this framework, by reducing the time and space required for computations.
In conclusion, we are offering a new framework for a more efficient addition of Transformers to the set of tools employed in 3D medical imaging, allowing even small labs working with heavy data loads to experiment at the forefront of deep learning technologies.

\section{Acknowledgments}
\begin{itemize}
    \item Funding: This work was supported by the STARS@UNIPD funding program of the University of Padova, Italy, through the project: MEDMAX.
    This project has received funding from the European Union’s Horizon Europe research and innovation programme under grant agreement no 101137074 - HEREDITARY.
    
    \item Code availability: The code to run our experiments will be available upon publication at: ...
    
    \item Data availability: The data for this study are accessible at: https://www.synapse.org/Synapse:syn28546456/wiki/633440
    
    \item Conflict of interest: The authors have no competing interests to declare that are relevant to the content of this article.
    
    \item LFT: conceptualization, investigation, writing - original draft preparation; AZ: formal analysis, writing - review \& editing; FDP: validation, writing - review \& editing; MA: resources, supervision, writing - review \& editing.
    
\end{itemize}

\newpage
\printbibliography
\end{document}